   \documentclass[acmtog,authorversion]{acmart}
   
\AtBeginDocument{%
  \providecommand\BibTeX{{%
    \normalfont B\kern-0.5em{\scshape i\kern-0.25em b}\kern-0.8em\TeX}}}

\setcopyright{none}

\acmSubmissionID{***}

\setcitestyle{nosort,square}

\begin{document}


\title{MonSter: Awakening the Mono in Stereo}

\author{Yotam Gil}
\email{yotamgil@mail.tau.ac.il}
\orcid{1234-5678-9012}
\affiliation{%
  \institution{Tel-Aviv University}
  \city{Tel-Aviv}
  \state{Israel}
}

\author{Shay Elmalem}
\email{shay.elmalem@gmail.com }
\orcid{1234-5678-9012}
\affiliation{%
  \institution{Tel-Aviv University}
  \city{Tel-Aviv}
  \state{Israel}
}

\author{Harel Haim}
\email{harel.haim.hh@gmail.com}
\orcid{1234-5678-9012}
\affiliation{%
  \institution{Toronto University}
  \city{Toronto}
  \state{Canada}
}

\author{Emanuel Marom}
\email{emanuel marom@tauex.tau.ac.il}
\orcid{1234-5678-9012}
\affiliation{%
  \institution{Tel-Aviv University}
  \city{Tel-Aviv}
  \state{Israel}
}

\author{Raja Giryes}
\email{raja@tauex.tau.ac.il}
\orcid{1234-5678-9012}
\affiliation{%
  \institution{Tel-Aviv University}
  \city{Tel-Aviv}
  \state{Israel}
}

\renewcommand{\shortauthors}{Gil, et al.}

\begin{abstract}

Passive depth estimation is among the most long-studied fields in computer vision. The most common methods for passive depth estimation are either a stereo or a monocular system. Using the former requires an accurate calibration process, and has a limited effective range. The latter, which does not require extrinsic calibration but generally achieves inferior depth accuracy, can be tuned to achieve better results in part of the depth range. In this work, we suggest combining the two frameworks. We propose a two-camera system, in which the cameras are used jointly to extract a stereo depth and individually to provide a monocular depth from each camera. The combination of these depth maps leads to more accurate depth estimation. Moreover, enforcing consistency between the extracted maps leads to a novel online self-calibration strategy.
We present a prototype camera that demonstrates the benefits of the proposed combination, for both self-calibration and depth reconstruction in real-world scenes.

\end{abstract}



\begin{teaserfigure}
  \centering
  \includegraphics[width=0.95\linewidth]{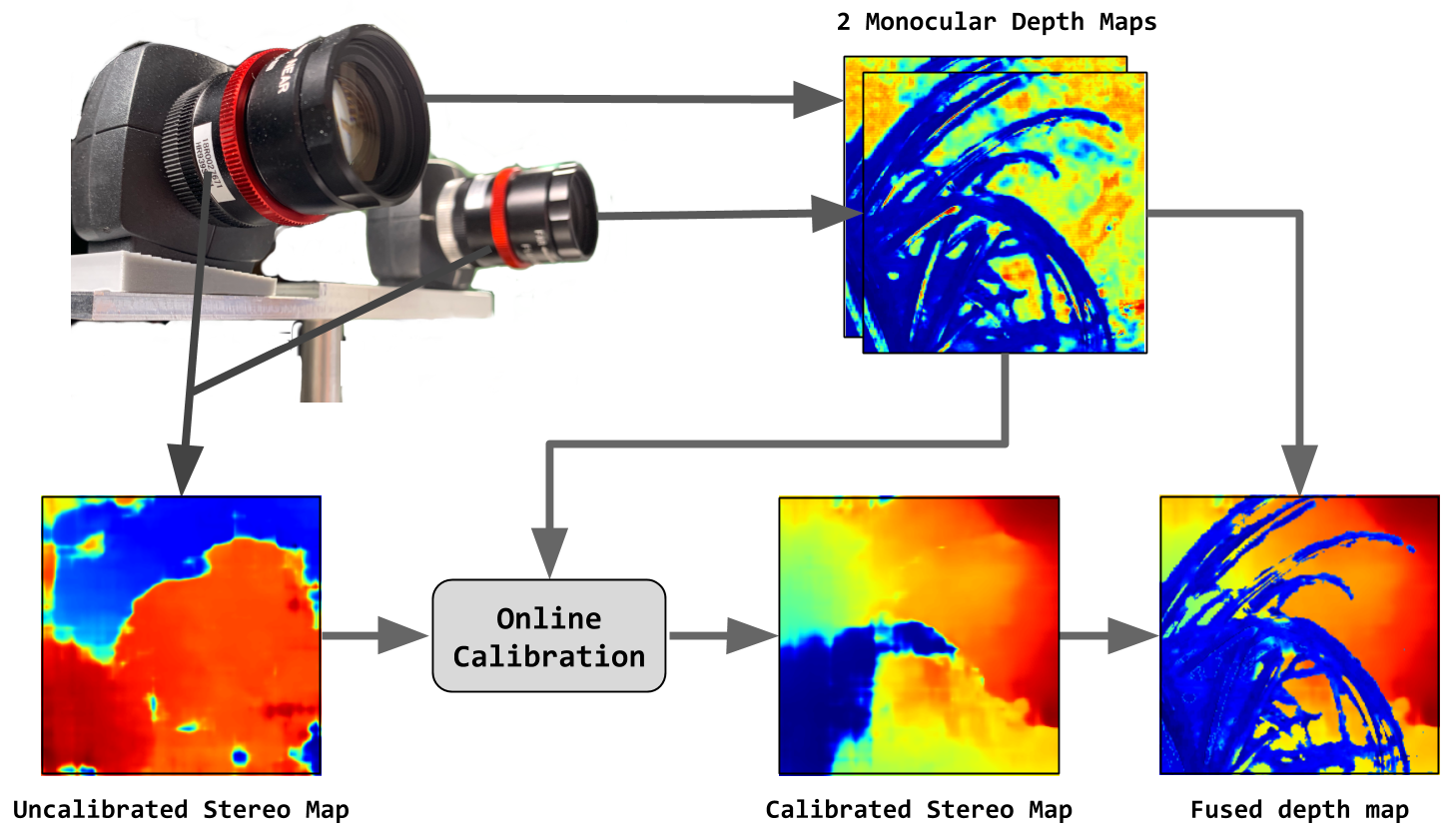}`
  \caption{Our mono-stereo system provides an extended range in depth estimation and an effective online calibration. Images acquired in a non-calibrated system provide a wrong depth map. By combining stereo and monocular techniques, our solution performs an automatic calibration that is better than other online schemes. Moreover, it improves the overall depth reconstruction mitigating some deficiencies in stereo (close-range errors) and mono (noisy far range).}
  \Description{Depth map improvement}
  \label{fig:real_world_improvement}
\end{teaserfigure}

\maketitle


\section{Introduction} \label{sec:intro}


\begin{figure*}
  \includegraphics[width=\textwidth]{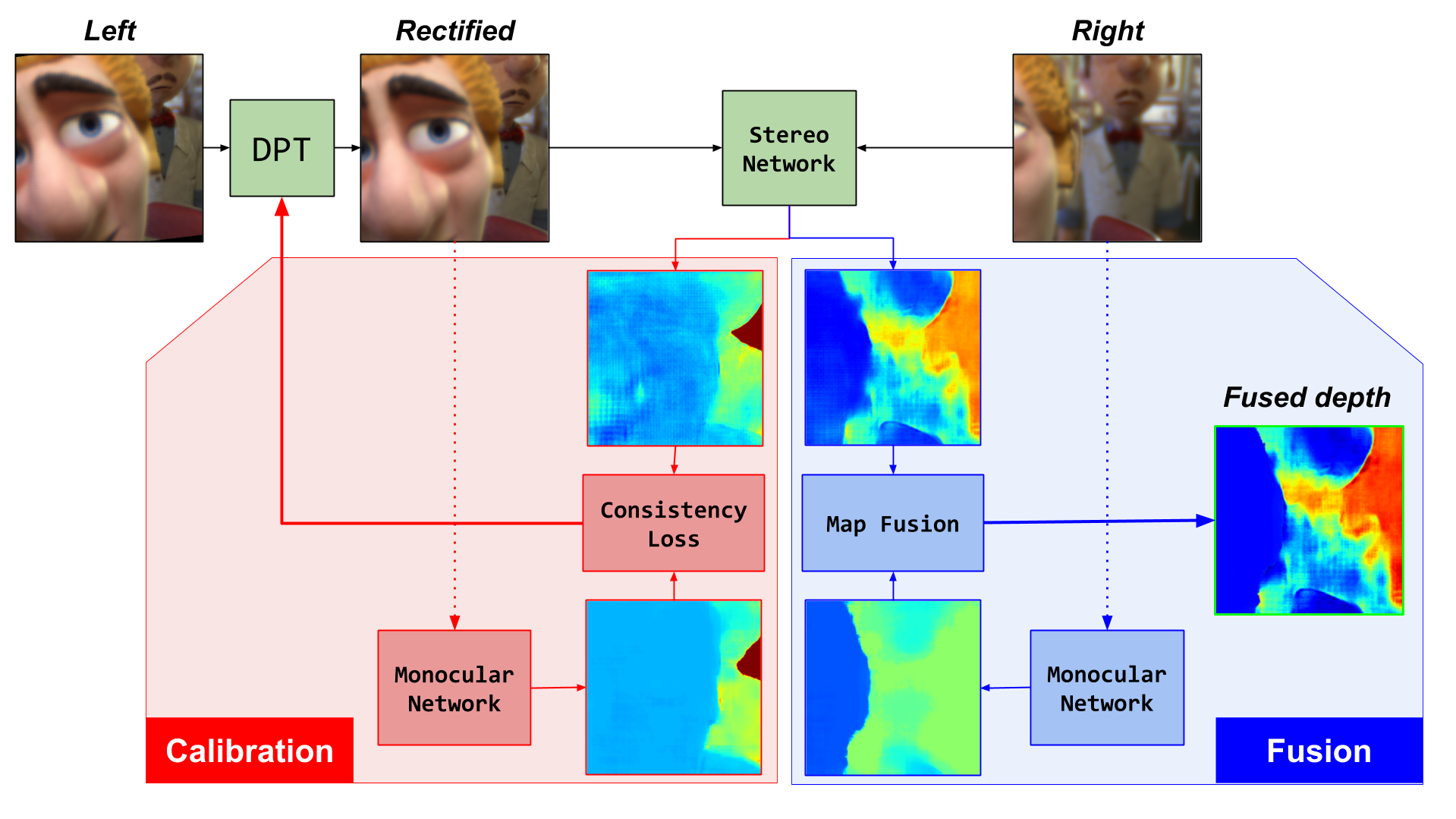}
  \caption{The proposed system presented on a simulation example. The left image is fed to a Differentiable Projective Transformation (DPT) and rectified to the right image. The rectified image and the right image are then processed in both the stereo and the monocular networks. For calibration (the red marked blocks), the system learns the projective transformation which provides the best consistency between the monocular and stereo left depth maps. For depth map fusion (the blue marked blocks), the right depth maps of the stereo and monocular images are fused into a more accurate depth map with an extended range.}
  \Description{Suggested System.}
  \label{fig:suggested_system}
\end{figure*}


Depth estimation is an important and challenging problem in computer vision and graphics.
Active methods for depth extraction include structured light (projecting known patterns onto the scene) or Time-of-Flight (ToF), which uses fast detectors to measure the phase shift between a reference signal and the returning signal from the object (using a continuous-wave laser) or to calculate the travel time of such a signal (emitted by a pulse laser). 

Passive methods for depth estimation rely on image information such as disparity (e.g., in stereo and light field cameras) or depth from perspective, shadows, focus/ defocus and more (mainly in monocular cameras). 
While active methods achieve more accurate results, they consume more power and generally require complex and expensive hardware, a complicated calibration process and achieve relatively low spatial resolution. While passive methods are usually based on cheaper hardware, they require higher computational efforts and achieve less accurate results compared to active methods.

In this paper, we propose an improved stereo imaging framework, which combines a conventional stereo vision method with two methods to extract monocular depth: a monocular phase-coded aperture technique, and an image-based monocular technique. Stereo vision aims at finding correspondence between two rectified images in order to estimate the disparity map between these two images. To rectify them, one is required to perform a supervised calibration process, which usually involves capturing several images of a known calibration pattern (such as a checkerboard target). If the physical structure of the camera has changed due to an intentional or accidental movement of one of the cameras, a re-calibration process is required. In addition, stereo based depth estimation methods are highly sensitive to occlusions and struggle to estimate depth in the proximate range, due to large disparities.

Monocular depth estimation aims at finding depth cues, which can be either global (such as perspective and shadows) or local (focus/out-of-focus). In this paper, we use and compare two monocular depth extraction methods:

1. An image-based monocular method, that estimates scene's depth based on the scene structure alone, and trained on many monocular depth datasets to improve its generalization ability \cite{DBLP:journals/corr/abs-1907-01341}. With recent years' advancements, a trained network can give appealing depth results based on one image - however, the depth map is relative and not in absolute distance units.

2. A monocular depth estimation based on a learned phase coded mask, presented in \cite{IEEE_depth}, which embeds depth related cues in an acquired optical image. These cues are in turn extracted by a Convolutional Neural Network (CNN), trained to estimate the scene depth according to those cues and enable a more accurate monocular depth estimation. The accuracy of phase-encoded monocular depth estimation depends on the ability to detect a change in focus for different depth ranges, thus, it is most effective around its focus point. Knowing the focus point of the camera, one can obtain an absolute depth map, with real-world distance units.

A fusion of both stereo and monocular methods can compensate for each method disadvantages and improve the overall depth estimation. 

First, since the monocular depth estimation does not require any calibration, it can be used as a reference for calibrating the stereo camera. By using such a reference, an online self-calibration process can be achieved by requiring consistency between the stereo and mono depth maps.

Second, since each method performs better on a different range (the stereo method works best for mid and far ranges, while the mono shows an advantage in close ranges), combining the two produces an extended depth range with improved accuracy. The second application is applicable for the phase-coded depth estimation only, as it allows setting the depth range in which it best performs by changing the focus point.


We demonstrate the performance of the system quantitatively on simulation data with ground truth depth maps. In addition, we show its advantage qualitatively on an experimental stereo setup that includes the phase masks. Our results show the advantage of combining the two methods over using stereo only for depth recovery. In the calibration side, we demonstrate our method using both phase-coded mask and image-based monocular method. we show better results compared to a conventional online calibration technique, and comparable performance to a checkerboard-based calibration, without the need for calibration targets imaging. 


{\em To summarize}, this work proposes a passive depth estimation imaging system, with the following important contributions:
\begin{itemize}
    \item A self-online calibration in a semi-supervised manner. 
    \item A proof-of-concept prototype camera and a demonstration of its performance in a real-world environment.  
    \item A method to combine monocular and stereo vision depth maps for achieving superior depth estimation. 
\end{itemize}

The rest of the paper is organized as follows. Section~\ref{sec:prevWork} reviews previous related work in passive depth estimation. Section~\ref{sec:method} describes the proposed method, and Section~\ref{sec:exp} presents both simulation and real-world experimental results. Section~\ref{sec:summ} concludes the paper and suggests possible future directions.

\section{Related Work} \label{sec:prevWork}

Passive depth estimation is a well-known challenge in computer vision and computer graphics, and extensive research has been performed in this area.
While many passive depth estimation techniques exist, in this work we focus mainly on stereo depth estimation, which is the first and most known approach in stereo vision, and monocular depth estimation using either aperture phase-coding mask or a using image features alone.

\subsection{Stereo Depth Estimation}
\label{sec:stereo_method}

Stereo vision works in similarity to the depth perception of the human visual system.
It uses two points of view to estimate the depth at each pixel by finding the disparity - the horizontal displacement of each pixel between the two acquired stereo images.
The disparity in location of the same object between two different images serves as a strong indication of the object's depth. 
The reconstructed depth's dynamic range and resolution are set together by the distance between the two cameras (known as the baseline), the cameras' field of view and the ability to accurately estimate the disparity. 

Recently, as in many fields of signal processing, deep learning (DL) is being used as the main tool to achieve improved stereo depth estimation results, starting from the initial patch-based works \cite{Zbontar2015ComputingTS,Luo2016EfficientDL}, to the more recent fully convolutional methods \cite{chang2018pyramid,zhang2019GANet,Yin2019,Cheng2019,Du2019}.
 
One should note that all of the known stereo algorithms assume that both images are rectified, which allows performing the search for the disparity only on horizontal lines. The transformations required for the images' rectification are achieved using a calibration process, which generally requires taking several images of a known calibration pattern (like a checkerboard target), making the process relatively tedious. In addition, the depth estimation performance is highly sensitive to calibration errors. Thus, a stereo camera design has high sensitivity to various environmental conditions (mechanical shock, vibration, thermal expansion) that can potentially change the setup calibration. Furthermore, in order to maintain the factory-made calibration, many of the stereo cameras sets are hardened - a fact that dictates baseline constraints, that can be avoided using online calibration. 

Several attempts were made to spare the need for a calibration object, by suggesting some online self-calibration techniques. Yet, they only showed limited success \cite{8PtsAlg,StereoFromUnCalib_Hartley,8PtsDefence}. Recently, DL based methods were utilized for multi-modal sensor registration \cite{RegNet,CalibNet} and for stereo calibration in planar roads scenario \cite{planarRoadCalib}. However, to the best of our knowledge, no generic self-calibration ability with comparable results to conventional checkerboard calibration was proposed.
 

In addition to the calibration requirement, there are few inherent pitfalls in the stereo method. The main and most-known issues are:
\begin{itemize}
    \item Out-of-frame and occluded pixels (pixels that appear only in one of the images). 
    \item A limited depth range - Finding proximate objects requires searching in a large disparity space, and can reduce the global depth accuracy.
    On the other hand, far objects will have a very small disparity and will, therefore, make differentiation in the far ranges difficult.  
    \item Repeated patterns on the horizontal axis challenge the finding of correspondences between the images pair. 
\end{itemize}

\subsection{Monocular Depth Estimation}

As the stereo camera approach is relatively complex, monocular depth estimation solutions have also been explored. Various monocular techniques use the global structure of the scene and depth cues like proportion and vanishing lines to achieve depth estimation, either in a supervised \cite{NYUD_1,NYUD_2,Make3D,depthMulScal_EigenFergus,DCNF_Liu} or an unsupervised \cite{Reid_GeometryRescue,depthLR} learning approach (for a survey on these methods and similar ones, see \cite{Bhoi2019_MonocularSurv}). Such methods achieve only relative depth estimation, and generally with limited performance. Moreover, learning based-methods that rely on global depth cues have limited generalization ability for scenarios different than the dataset used for training. The method we use in this paper, based on \cite{DBLP:journals/corr/abs-1907-01341} attempts to tackle the generalization limitation by training on many different aligned datasets. 

Other monocular solutions use optical cues. Since the lens response is depth dependent (due to different behaviour of in- and out-of-focus), this feature can be employed for depth estimation. Under this category one may find either depth from focus/defocus \cite{DFF_darrel,DFF_Schechner2000,Trouve:13,Suwajanakorn_2015_CVPR_DffPhone,Carvalho_2018_ECCV_Workshops,Wolf_DFD}, and depth from a focal stack \cite{Coded_DFFS_Raskar,Lin_DFFS,hazirbas18ddff}.
A recent work \cite{doubleLytroPaper} attempts to combine two focal stacks (acquired using a light-field stereo pair) to achieve improved depth estimation.

A more sophisticated approach employs computational imaging - a method in which a modification is done to the imaging system in order to acquire an optical image that better suits the final application \cite{compImg_rev}. In the case of depth estimation, by coding the lens response in a certain way, the depth-dependent behaviour of the optics is intensified, such that the optical depth cues embedded in the image are much stronger. Such methods have been presented for both amplitude aperture coding \cite{Levin_coded_aper,Nayar_codAperPair,mitra2017CodedAperture},  phase aperture coding \cite{Nayar_depthFromDiff,IEEE_depth,wu2019phasecam3d,WetzsteinDeepOpt} and spectral aperture coding \cite{chakrabarti2012depth,Martinello2015DualAperture}. 

As previously mentioned, this paper will compare and apply both monocular approaches to present self-calibration ability, and the phase-coded method to improve overall depth-map estimation. 
Since the phase-coded method is less known and straightforward, we hereby present its basic idea and principles.

\subsection{Monocular Depth Estimation Using Aperture Phase-Coding}

We focus on aperture phase-coding, which has the advantage of having a very-high light efficiency, with almost no loss. In particular, we use the technique proposed by \citet{IEEE_depth}. They suggest using an aperture phase-coded mask in the image acquisition process. Using the phase-coded mask, a depth and color dependent point spread function (PSF) is generated, such that each of the image's RGB channels can be thought of as being optimally focused on a different depth in the scene. 

Using the focus and out-of-focus color-depth cues embedded in the image, a neural network can be used to predict the defocus condition (labeled as $\psi$) at each pixel. Assuming the lens parameters and focus point(s) are known, the absolute depth can be easily derived from $\psi$, which is given by
\begin{eqnarray*}
\psi = \frac{\pi R^2}{\lambda}(\frac{1}{z_o} + \frac{1}{z_{img}} - \frac{1}{f}) = \frac{\pi R^2}{\lambda}(\frac{1}{z_{img}} - \frac{1}{z_i}) = \frac{\pi R^2}{\lambda}(\frac{1}{z_o} - \frac{1}{z_n}),
\end{eqnarray*}
where $R$ is the radius of the exit pupil (assuming a circular aperture), $\lambda$ is the illumination wavelength, $z_{img}$ is the sensor plane location for an object in the nominal position $z_n$, and $z_i$ is the ideal image plane location for an object located at $z_o$. As $\mid{\psi}\mid$ increases, the image contrast decreases, hence the contrast is at maximum for $\psi=0$ (the in-focus position). The mask and method presented in \cite{IEEE_depth} are designed for optimal depth estimation in the range of $\psi=-4..10$.


\section{Method}
\label{sec:method}
   
\begin{figure}
  \centering
  \includegraphics[width=\linewidth]{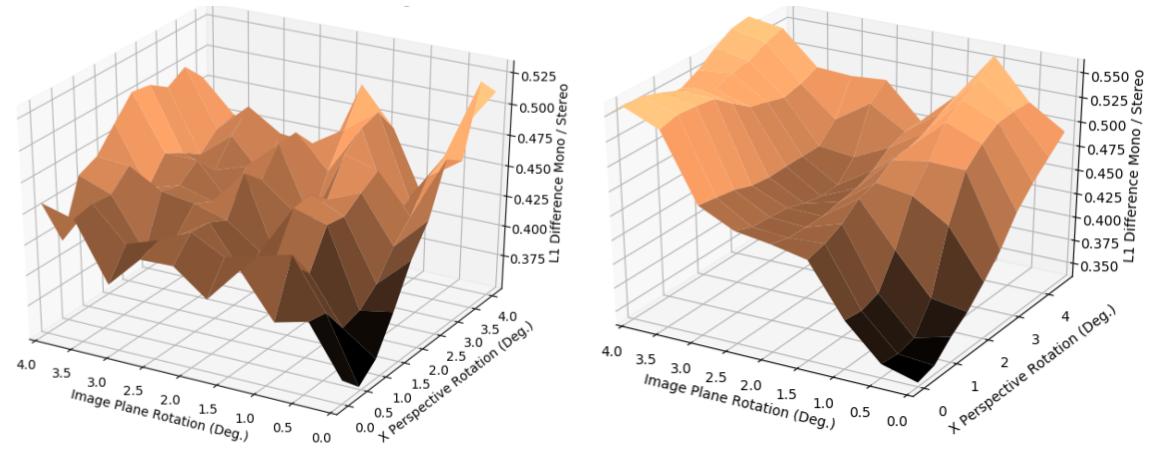}
  \caption{The L1 difference between the mono and stereo depth maps when rotating the right image over two axis: the image plane and the $X$-perspective, for two monocular methods: Image-based (Left) and phase-coded (Right). Notice that in both methods, the minimal difference is achieved when the stereo setup is calibrated. The difference in the phase-coded method is smoother, hence less prone to get a local-minima calibration rather than the real one.}
  \Description{L1-Difference for rotation}
  \label{fig:diff_for_rotation}
\end{figure}

Acknowledging the strengths and weaknesses of the stereo and monocular methods enables combining them efficiently, and gain from the advantages of each method. 
Thus, our proposed approach integrates the stereo and monocular depth estimation techniques to achieve two goals:
\begin{itemize}
    \item Self-calibration abilities of a stereo vision setup (see Figure~\ref{fig:suggested_system}(left)). 
    \item Improvement of the overall depth estimation by fusion of the two depth maps (see Figure~\ref{fig:suggested_system}(right)). 
\end{itemize}

In order to demonstrate our phase-coded masks system, we offer the use of two identical phase-encoded masks incorporated in both lenses of the stereo camera, while each camera is focused on a different depth. The right camera is focused on a closer focus point (we use $0.7[m]$ in our experiments) to improve the inherent limitation of stereo vision in the proximate depth ranges. The left camera is focused on a farther range (we use $1.5[m]$ in our experiments) and is used to achieve a monocular depth map covering a broader range of depths, to serve as a reference for the self-calibration process. As described in the previous section, the estimated $\psi$ parameter in the monocular method spreads a range of depths around the focus point and it is most accurate in its proximity.

To show the self-calibration process using the image-based monocular method, we will simply use the aforementioned stereo cameras set. Our experiments show that the phase-masks will not affect the quality of results for both stereo and image-based monocular methods, due to their global nature and ability to ignore the phase-masks local optical cues.

\subsection{Auto self-calibration}

As the monocular depth estimation has no requirement for an extrinsic calibration process, we can use it as a reliable source for self-calibration by requiring consistency between the monocular and stereo depth maps. The assumption is that consistency between the two sources of depth maps (mono and stereo) would be optimal when the stereo setup is calibrated, as the stereo depth map would be most accurate then. 

This assumption is empirically tested by examining the L1 difference between the stereo and mono depth maps for various calibration errors, as presented in Figure~\ref{fig:diff_for_rotation}. For the sake of visualization, we chose two rotation axes: image plane rotation and $x$-axis perspective rotation. We rotated the right image along the axes and calculated the L1-difference between the perceived mono and stereo depth maps. It can be seen that the minimal difference is achieved when both angles are zero, and that difference increases with rotation. As can be seen in the figure, the error in the phase-coded method is more smooth, hence more likely to find the real calibration. This finding is evident in our experiments, in which we saw that the image-based method is more sensitive to initialization.

Following the method suggested in \cite{NIPS2015_5854}, we incorporate a differentiable projective transformation block with 8 degrees of freedom, applied on either the left or the right image in the input stage of the stereo network. This block can learn the required transformation parameters to rectify this image to the other image. In our experiments, we also tried using two differentiable projective transformation blocks (for both the right and left images) in order to rectify both images to a common plane, however, we witnessed inferior results supposedly due to an increased number of solutions. For the auto-calibration process, both the weights of the pre-trained stereo depth network and the pre-trained monocular depth network are frozen, and only the projective transformation parameters are trained in order to rectify the images in the non-calibrated stereo setup. This method can be considered semi-supervised, as it needs no ground truth of the real depth of the scene, but it is using pre-trained networks that were previously trained with depth ground truth. 

\begin{figure}
  \centering
  \includegraphics[width=\linewidth]{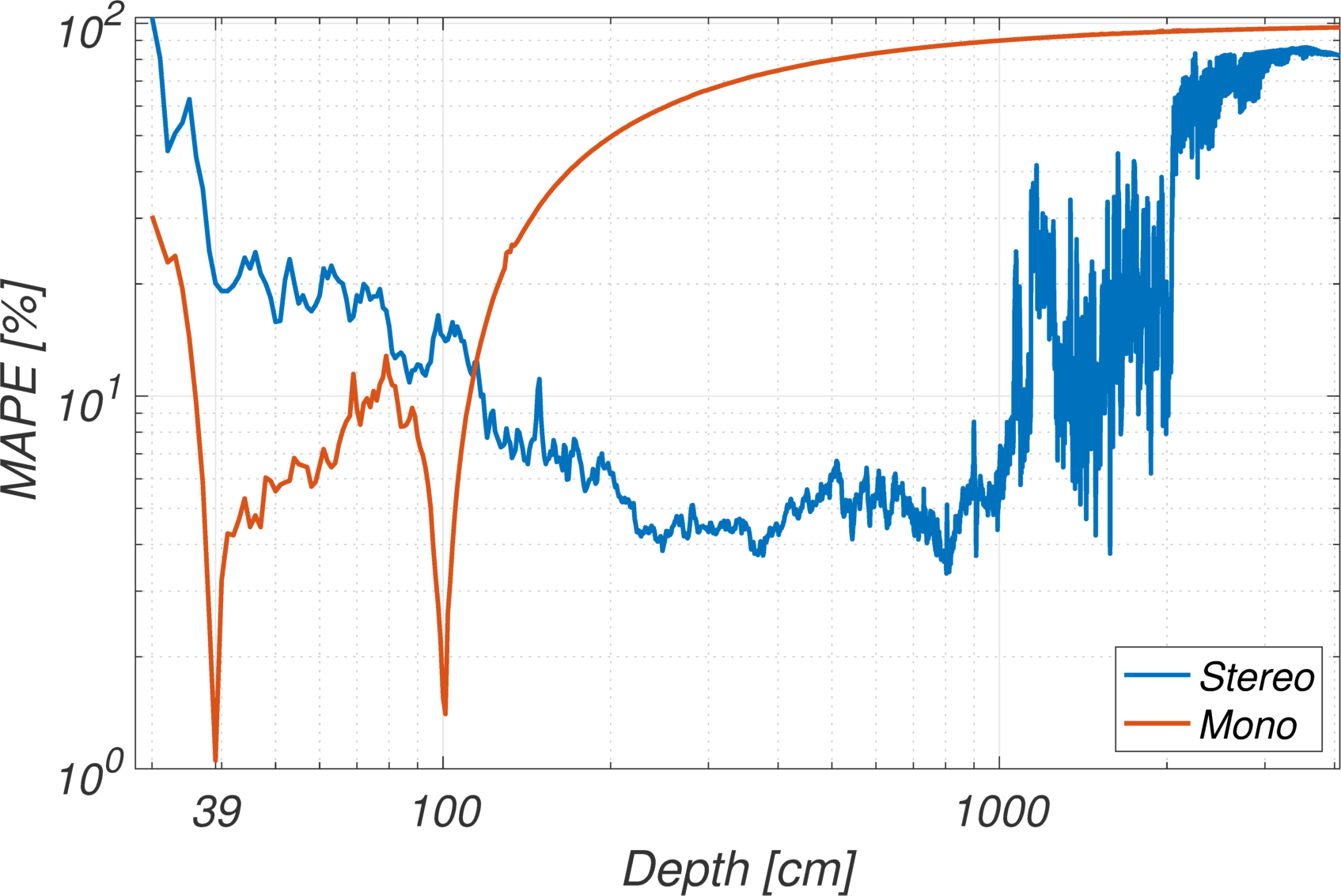}
  \caption{Mean absolute percentage error of Stereo with a $10 [cm]$ baseline and phase-coded mono, focused at $0.7[m]$, on a large labeled dataset. The monocular method shows superior depth estimation in the range of $0.39-1.0[m]$, while the stereo shows superior depth estimation at the farther distances.}
  \Description{Relative loss of Stereo and Mono.}
  \label{fig:rel_loss}
\end{figure}

Using the phase-coded method for calibration would require a scene in a specific depth range: an optimal image for calibration is an image that most of its pixels are within the depth range spread by $\psi=-4..10$. In this case, there will be a maximal amount of features for comparison. In our examples of a focal point of $1.5[m]$, the range spread by $\psi=-4..10$ is $0.56-4.5[m]$. 

However, calibration using the image-based method is possible for every depth range the monocular network was trained upon. 

The auto-calibration process may be performed in two ways:
\begin{itemize}
    \item User-initiated calibration: the user takes a set of left and right image (for the phase-coded calibration, the images would be with the proper depth range that fits the $\psi$ parameter range), and then initiates the training process of the projective transformation parameters accordingly.
    \item An offline calibration: the system automatically chooses an image set from a library, with the most pixels in the monocular depth range, and fine-tunes the current calibration using these.
\end{itemize}

Another benefit of our approach is that it can indicate whether the system is out-of-calibration by noticing a decrease in the monocular and stereo depth map consistency and thus, alerting the user so he or she can initiate a calibration process.

\subsection{Improving depth estimation}
As mentioned in Section \ref{sec:stereo_method}, stereo vision suffers from high error in proximate ranges due to large disparity values, where in terms of relative error, the error is even more prominent. Phase-encoded monocular depth estimation is most accurate around its focus point, thus by setting its focus point to a close depth we can use its depth estimation to improve the stereo depth estimation. In this section we present the depth-map enhancement using phase-coded method only, and not image-based monocular method, since image-based method provides a relative depth estimation only. Even if some alignment to an absolute metric is done, our experiments show that the stereo method will be more accurate throughout the entire depth range, hence cannot be improved using the image-based monocular method.

As Figure~\ref{fig:rel_loss} shows (we provide more details on this figure hereafter), we can use a phase-coded monocular approach to improve depth estimation in the close ranges. In addition, this allows to decrease the maximal disparity search space of the stereo method, and by that improve stereo depth estimation accuracy in other ranges. 
The integration between the monocular and the stereo depth maps is done with a fusion network, that merges the depth maps according to the confidence of each method in its depth prediction. 

As shown in Figure~\ref{fig:suggested_system} (right), the depth reconstruction is performed by fusing the output of two neural networks, one for stereo depth estimation based on \cite{chang2018pyramid} and one for monocular depth estimation using \cite{IEEE_depth}. 

The stereo network based on \cite{chang2018pyramid} and the image-based method based on \cite{DBLP:journals/corr/abs-1907-01341} are trained to output the scene disparity.

The monocular network \citet{IEEE_depth} is trained to predict a $\psi$ value in the range of $[-4:10]$. Given a fixed aperture size $R = 1.14mm$, the camera focus point can be used to determine the depth dynamic range spread by the predicted $\psi$ range. In order to increase the monocular depth accuracy for the close range while keeping an overlap between the stereo and mono depth maps, we set the focus points for the right and left cameras at - $0.7m$ and $1.5m$, that spread the range of possible predicted distances to $0.39-1 [m]$ and $0.56-4.5 [m]$ respectively. 
Some of our experiments were done with disparity and some with depth. 
In order to estimate the monocular versus stereo depth estimation performance, a large set of 500 pairs of stereo labeled images is used to estimate the error in both methods. The results are presented in Figure~\ref{fig:rel_loss}, in Mean Absolute Percentage Error (MAPE). One can note that the monocular camera focused at $0.7[m]$ achieves superior depth estimation in the range of $0.39-1.0[m]$ over the stereo method with a baseline of $10[cm]$. Thus, the monocular depth estimation can help improving the stereo depth map in this proximate range. The monocular depth map acquired from the second camera (containing information of a broader depth range) is used as a built-in reference for calibrating the stereo setup, as discussed in the previous subsection.


To fuse the stereo and monocular depth maps, both the stereo and the monocular networks are enhanced with an additional confidence 'head' before their last regression layer that provides their output. 
The confidence values of the two networks are then used to create a binary mask that determines which depth map should be used for each pixel in the fused depth map. The overall network is trained in an end-to-end fashion with the depth ground truth, using a smooth L1-loss and the Adam optimizer.

\section{Experiments} \label{sec:exp}
   The proposed system and both tasks of auto-calibration and depth estimation enhancement were initially tested using simulated images. Scenes containing both high-quality RGB images and their corresponding depth maps were created using the Blender software. 

The dataset consists of 500 pairs of rectified stereo images (with a baseline of $10 [cm]$) and their ground truth depth maps \footnote{The dataset will be made publicly available upon publication.}. A proper imaging simulation process was applied to the images, modeling the phase-encoded mask and depth-dependent imaging effects. A stereo network based on \cite{chang2018pyramid} was trained, following its success in the KITTI2012 and KITTI2015 challenge. This network shown great results using atrous convolution to exploit scene's global context. We extended the stereo network to output a disparity map for both the left and right images. We also trained two types of networks for the monocular depth estimation: 

\begin{itemize}
    \item A phase-encoded monocular network, based on \cite{IEEE_depth} using the created dataset. This network is a 5-stages fully-convolutional network. The network is relatively shallow, as it only needs to find local de-focus cues rather than understanding a global context. 

    \item An image-based monocular network, based on \cite{DBLP:journals/corr/abs-1907-01341}. This network novelty was it being trained on many different monocular depth datasets, thus overcoming the generalization problem attributed to other image-based monocular methods. 
\end{itemize}

\begin{figure}
  \centering
  \includegraphics[width=\linewidth]{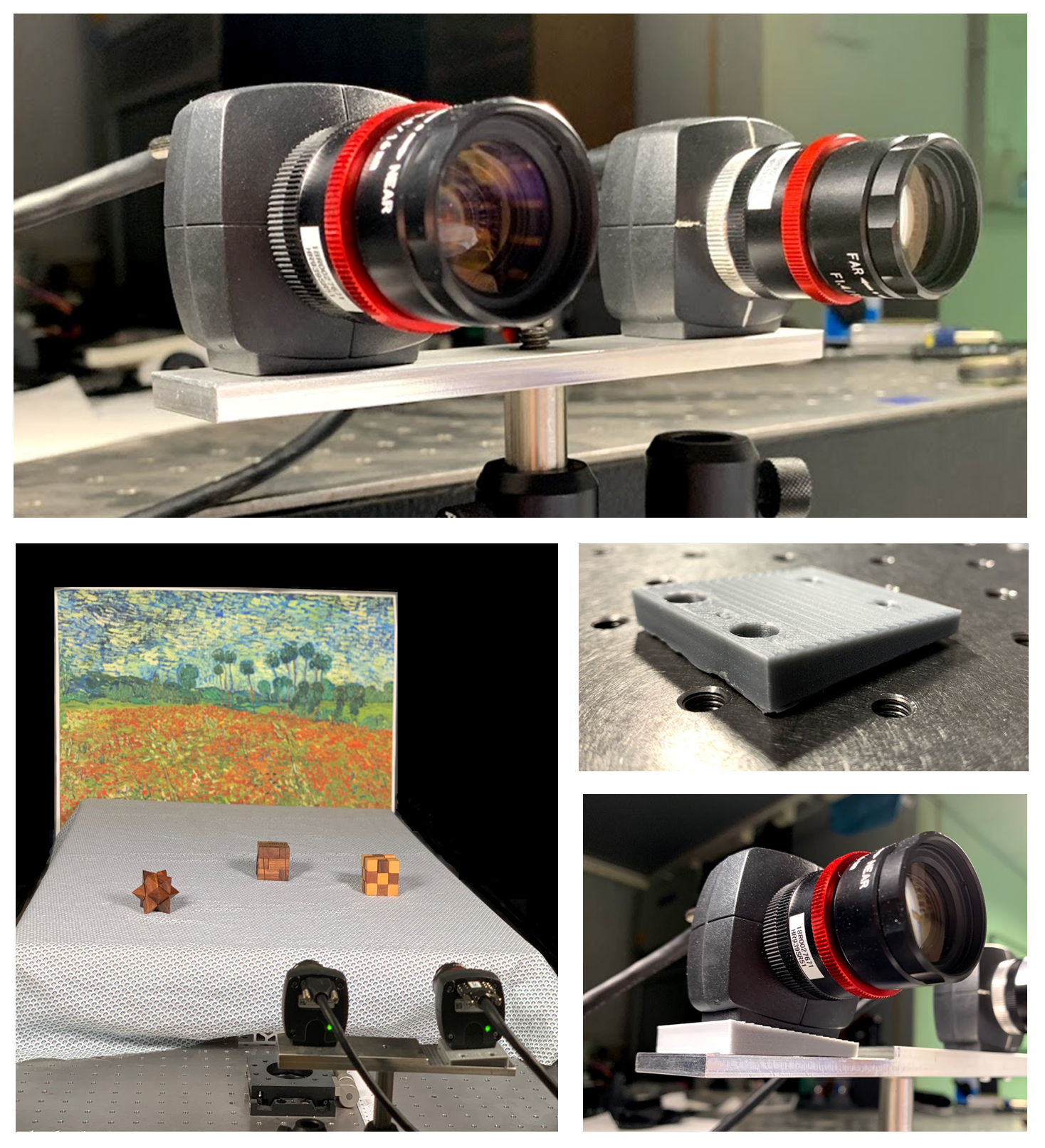}
  \caption{Our prototype. Top: the stereo set, with a phase-encoded mask applied on each camera. Bottom left: An indoor test scene example. Bottom right: An additional 5 degrees image plane rotation is applied to the right camera. It allows testing our calibration method on such deviations.}
  \Description{our system}
  \label{fig:system_5deg_rotation}
\end{figure}

After achieving satisfying results on the simulated images, a real-world experiment was performed. The experimental setup is based on two IDS3590 18MP cameras equipped with KOWA LM16JCM-V lenses (with $f=16[mm]$ focal length) and proper aperture-phase masks (see Figure~\ref{fig:system_5deg_rotation}). The cameras are mounted as a stereo pair with a $10 [cm]$ baseline. Note that the real-world images are processed using the models trained on the simulated images and with no fine-tuning on real scenes. The results show that the method generalizes well to real-world images, which is a strong indication for its robustness.

\subsection{Auto self-calibration}
The auto-calibration process is presented in Figure~\ref{fig:suggested_system}. As the simulation images are rectified, calibration error was simulated by transforming the left images in our dataset using an arbitrary projective transformation. We then trained a network to learn the specific projective transformation that achieves monocular-stereo depth consistency. The pre-trained stereo and mono networks weights were kept frozen, and only the projective transformation parameters were trained with back-propagation through the stereo network. For the phase-coded method, The focus point used for the left camera was $1.5[m]$, that makes the $\psi=-4..10$ parameter spread depth in the range of $0.56-4.5[m]$ so that the monocular depth map is accurate in these ranges. The loss used is the L1-loss, an absolute distance between the stereo depth map and the monocular depth map. 

We next compare L1 and relative-L1 difference between the stereo depth map and ground-truth depth map, after rotating the images in an arbitrary rotation of 7-degrees in the image plane, and rectifying the images using 4 different methods - extracted SURF features following \cite{8PtsDefence}, our method with a phase-coded monocular method and projective transformation applied for both left and right images, our method with a phase-coded monocular method and projective transformation applied only for the left image, and our method with an image-based monocular method and projective transformation applied only for the left image (the 2 last options are as seen in Figure~\ref{fig:suggested_system}). 

The training of the calibration in our method is done with a pair of images for 100 epochs, and averaged over 10 training processes. The results can be seen on Table~\ref{table:cal_compare}. As can be seen, the results of our method using one differentiable projective transformation shows significant superior results over the two other calibration methods presented in the table.

\begin{table}
  \centering
  \caption{The L1 and relative-L1 difference between stereo depth maps and ground truth depth maps, after a calibration using three methods: SURF features extraction (~\cite{8PtsDefence}), rectifying the images using a differentiable projective transformation (DPT) for both images (Two DPT, using phase-coded method) and using DPT only for the left image, using both image-based and phase-coded methods.}
  
  \begin{tabular}{cccc}
    \toprule
    Method&L1-loss& Rel-L1-loss\\
    \midrule
    SURF features & 2.02 & 0.56 \\
    Two DPT (phase-coded) & 1.18 & 0.34\\
    One DPT (image-based)& 0.92 &  0.28\\
    One DPT (phase-coded)& \textbf{0.84} &  \textbf{0.22}\\
  \bottomrule
  \label{table:cal_compare}
\end{tabular}

\end{table}

\begin{figure*}
  \centering
  \includegraphics[width=\linewidth]{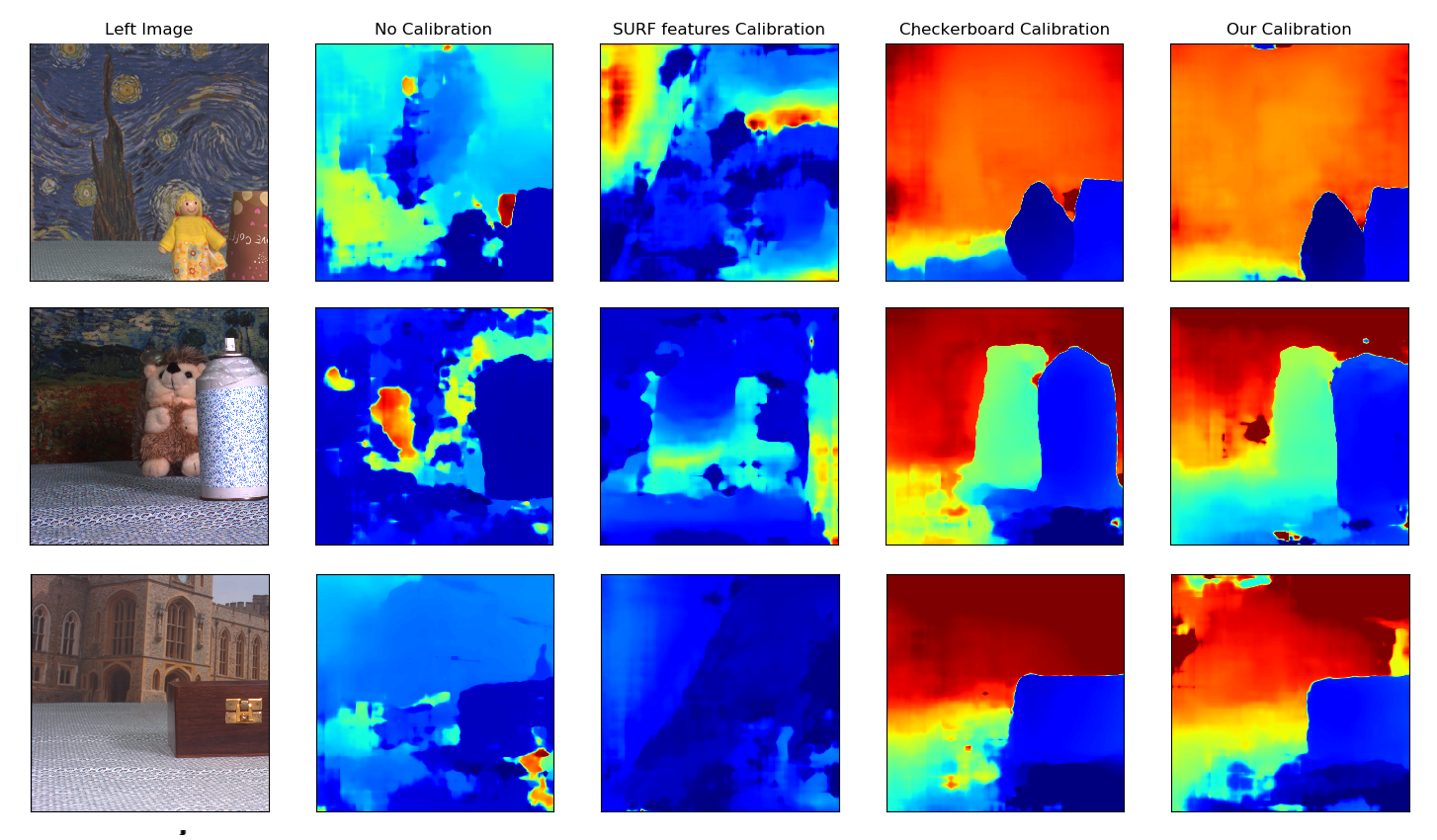}
  \caption{Examples of our online-calibration results on a real-world image. From left to right: the RGB left image and the depth results of the stereo network on a pair of images that is: non-calibrated, calibrated using a matching of extracted SURF features as in  ~\cite{8PtsDefence}, calibrated using a checkerboard and using our self-calibrating method. It can be seen that our technique is on par with the conventional checkerboard calibration method.}
  \Description{Results of different calibration methods}
  \label{fig:cal_comparison}
\end{figure*}

\begin{figure*}[h]
  \centering
  \includegraphics[width=\linewidth]{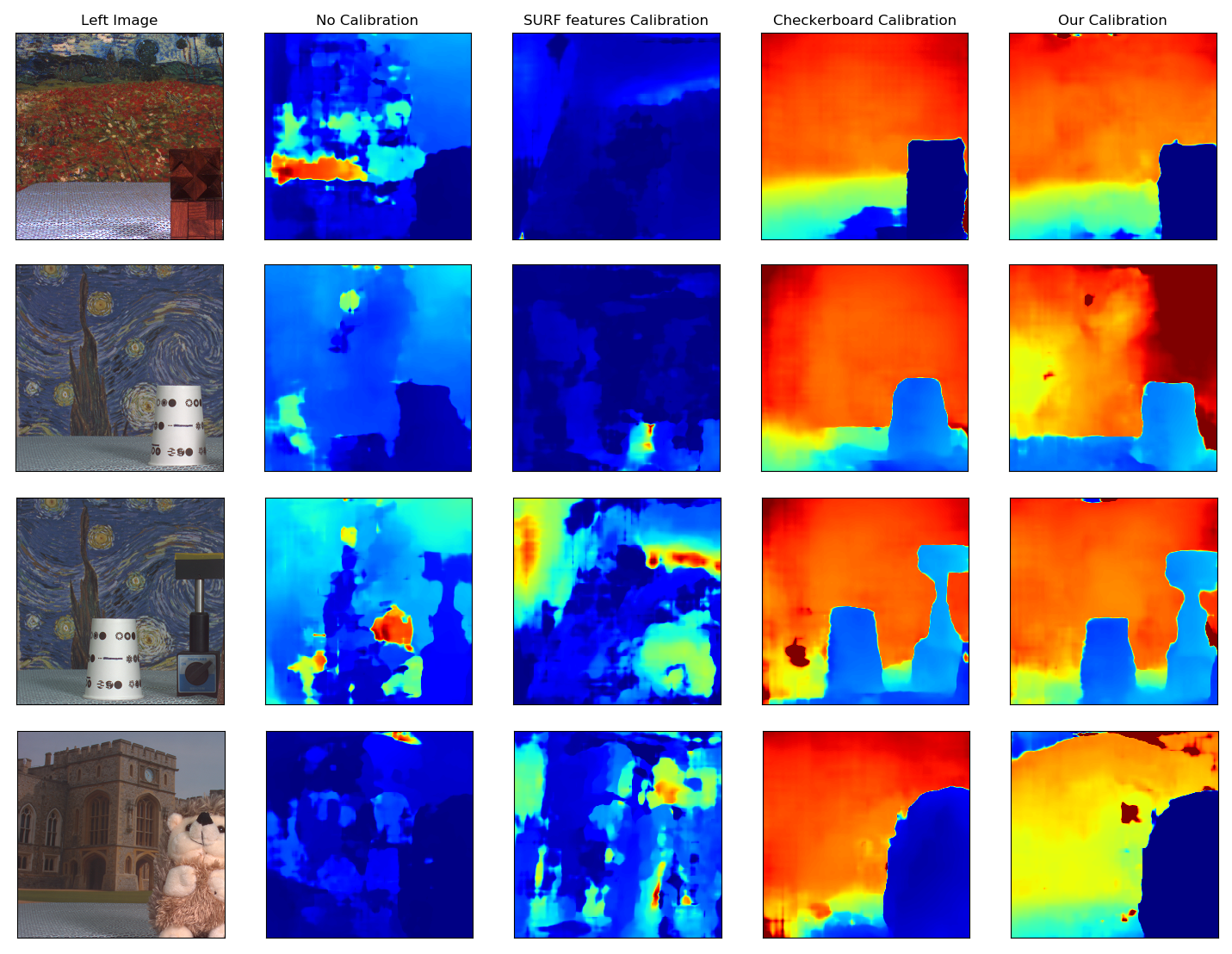}
  \caption{Depth estimation of images rotated with additional 5 degrees, after calibration with SURF features, checkerboard calibration and our method. Note that our approach is able to correct the relatively high deviation and calibrate the system correctly. Its resulted depth is on par with the checkerboard outcome. }
  \Description{our system with additional rotation}
  \label{fig:5deg_rotation}
\end{figure*}

Following the successful simulation results, we moved to real-world images. We first tested a calibrated setup and transformed it to simulate a known out-of-calibration state. Then we tested our system 'in the wild', on a non-calibrated setup. As shown in \cite{chang2018pyramid} and \cite{IEEE_depth}, the pre-trained stereo and monocular networks were able to adapt well from the simulated domain to the real-world domain. We trained the projective transformation parameters with the real-world images. As most of the network is pre-trained, and our training includes only the projection transformation parameters, the training process is fast - we only need one pair of images, and we train for about 100-200 epochs. The entire training takes less than 3 minutes on one Nvidia GTX2080Ti. 

In Figure~\ref{fig:cal_comparison}, we compare depth estimation results after calibration using 3 methods: a standard calibration process based on a known pattern (checkerboard target calibration), an image-based calibration process achieved by extracting image features using the SURF algorithm and rectifying the images according to these features \cite{8PtsDefence}, and our method. It can be seen that the SURF-based algorithm was not able to rectify the images properly, while our calibration method shows very similar results to the conventional calibration process - without the requirement of taking a known pattern calibration target. our calibrations were done using the image-based and phase-coded monocular methods, and achieve similar results.

\begin{figure*}[h]
  \centering
  \includegraphics[width=\linewidth]{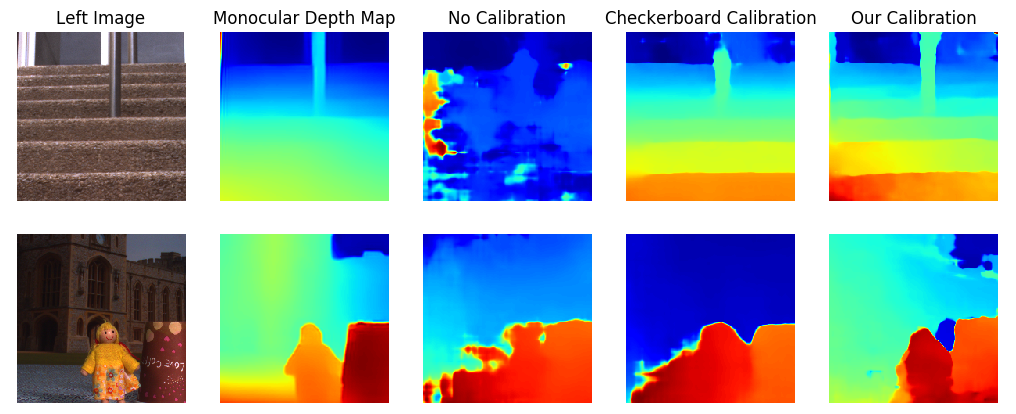}
  \caption{Auto-calibration examples using image-based monocular method. Using the monocular depth map as a reference, the system auto calibrates itself to achieve the rightmost result. In some cases, our method shows superior results over the checkerboard calibration - like the rod on the first row or objects structure in the second row. However, some of the monocular method error appears now on the stereo result - like the right top corner in the second row, while the background is completely flat. }
  \label{fig:stairs_example}
\end{figure*}

Figure~\ref{fig:stairs_example} shows another example of calibration of an outdoor staircase. This figure shows that our calibration method can even perform better than the conventional calibration method - the rod on the left side of the image appears only on the depth map of our calibration.

We also checked our calibration on a larger deviation of the stereo setup. One of the cameras in our non-calibrated setup was now rotated by 5 degrees in the image plane, as can be seen in figure \ref{fig:system_5deg_rotation}. As seen in Figure \ref{fig:5deg_rotation}, we were still able to self-calibrate our system and achieve comparable results to those of the checkerboard calibration.

\begin{figure*}[h]
  \centering
  \includegraphics[width=\linewidth]{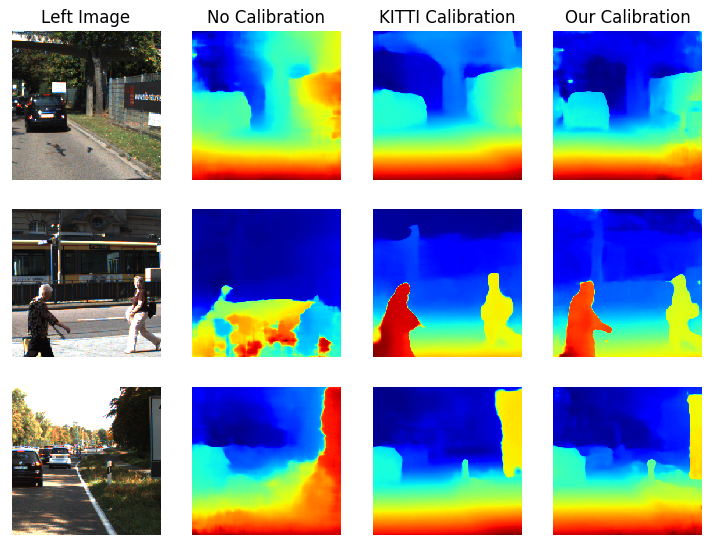}
  \caption{Auto-calibration on KITTI raw dataset. In some cases, our calibration is more accurate than the conventional method used to rectify the KITTI images.}
  \label{fig:KITTI_calib}
\end{figure*}

Our next experiment was to rectify KITTI images \cite{Geiger2013IJRR}. The raw KITTI dataset is available online, and we used it to demonstrate our calibration method and compare it to the calibration done by the authors of the dataset, using conventional checkerboard patterns. Since KITTI dataset includes scenes of roads and streets, the phase-coded method cannot be used here - the de-focus cues disappear at these far range distances. Hence, we used the image-based monocular method. We compare the L1 distance between the depth maps obtained using no calibration, our calibration and the conventional checkerboard calibration, to the rectified KITTI images ground-truth. Results can be seen in table~\ref{table:KITTI_calib}. Calibration examples can be seen in ~\ref{fig:KITTI_calib}.

\begin{table}
  \centering
  \caption{L1 average distance between depth maps obtained using no calibration, KITTI checkerboard calibration and our calibration, and KITTI ground truth.}
  
  \begin{tabular}{cc}
    \toprule
    Method&L1-loss\\
    \midrule
    No Calibration & 1.23  \\
    Checkerboard Calibration & 0.835 \\
    Our Calibration & 0.909 \\
  \bottomrule
  \label{table:KITTI_calib}
\end{tabular}

\end{table}

\begin{figure}
  \centering
  \includegraphics[width=\linewidth]{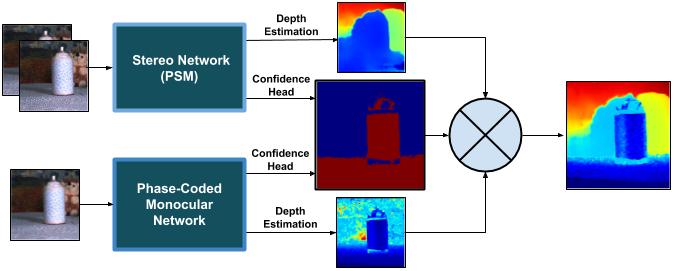}
  \caption{The Fusion network predicts a binary-mask (blue and red image) based on the confidence heads from the stereo and the monocular networks, to determine the source for each pixel in the fused depth map. }
  \Description{our system}
  \label{fig:fusion_net}
\end{figure}

\subsection{Improving depth estimation}
As mentioned in Section \ref{sec:stereo_method}, the stereo method shows inferior results in the proximate ranges, while phase-coded monocular depth estimation can be tuned to be most accurate on a desired specific range. Thus, we use a camera with a focus point of $0.7[m]$, so it would be most accurate in the ranges of $0.39-1[m]$, where the stereo method suffers from the largest relative error. As Figure~\ref{fig:rel_loss} shows, the monocular estimation is substantially better in this depth range. 

As described in Section~\ref{sec:method}, we used a neural network to combine the monocular and stereo depth maps. The network, based on the confidence of the mono and stereo networks predictions, outputs a a binary mask to determine the source of each pixel in the fused depth map (see figure \ref{fig:fusion_net}). 
To get a confidence metric, we concatenated the channels of each network before the regression layer (that multiplies each channel with its corresponding depth or disparity). The stereo and monocular networks have 192 and 16 channels correspondingly, so the input to the fusion network is of size (Bx(192+16)xWxH). We used a 4-layers' convolutional network, to get a binary prediction for each pixel in the fused depth map.


Table~\ref{table:fusion_loss} compares the fused depth to the monocular and stereo ones. The results are shown on simulated data that have ground truth depth maps. The map fusion shows an improvement of 10\% in the relative L1-loss (measured between the estimation and the depth ground truth).

\begin{figure*}
  \centering
  \includegraphics[width=\linewidth]{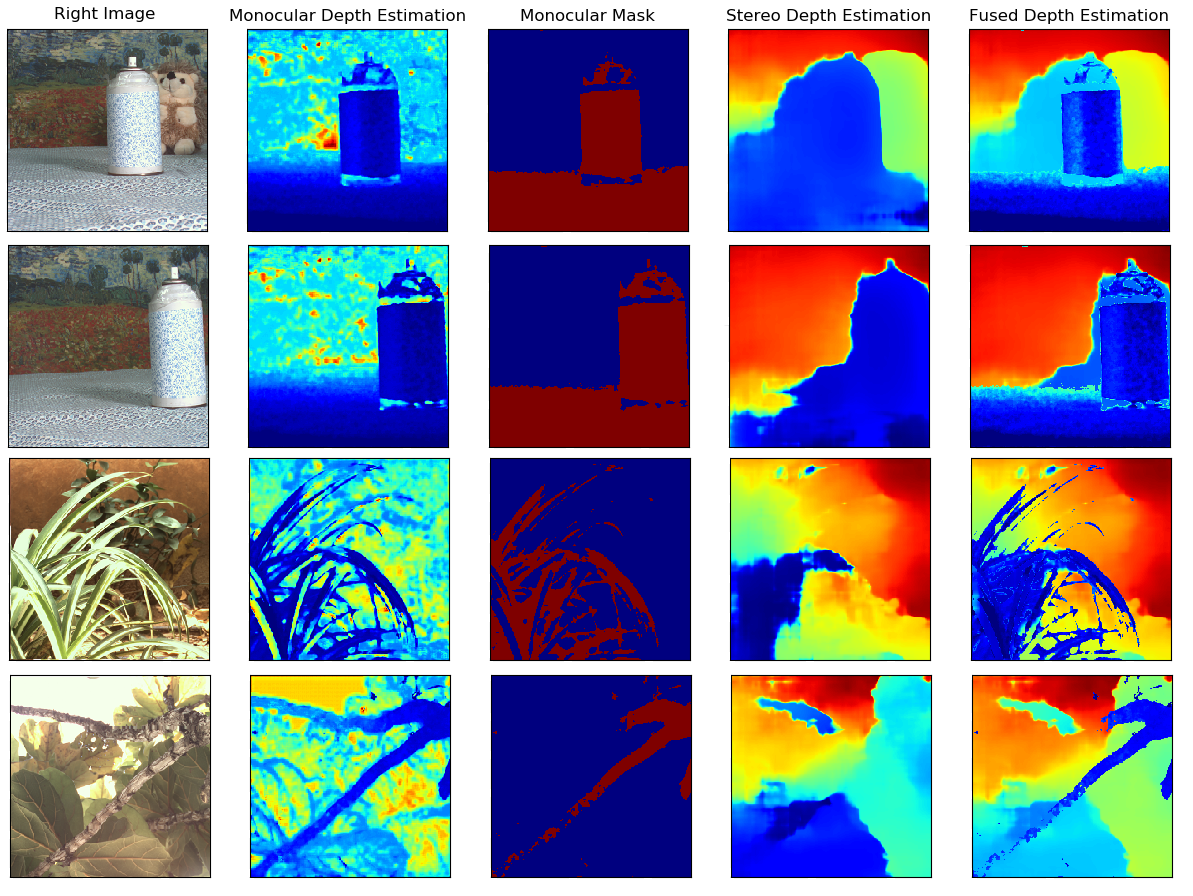}
  \caption{Examples of fusion of Stereo and Mono depth maps: the first two rows show examples of a table with a close spray-bottle on it. The monocular method is able to accurately estimate the gradual depth of the table and the bottle, while the stereo method estimates the background better. In the two last rows, the fused depth maps add objects from the monocular depth map that are not properly perceived in the stereo depth map due to their proximity.}
  \Description{Fusion Examples}
  \label{fig:fusion_examples}
\end{figure*}

Having superiority in the quantitative results, we turn to demonstrate the advantage of the approach qualitatively on real-world images. Figure~\ref{fig:fusion_examples} presents the results of our prototype. 

\begin{table}
  \centering
  \caption{Relative Loss of Different Methods. As can be seen in Figure \ref{fig:rel_loss}, the monocular method is better than the stereo in a specific depth range. Fusing depth estimation from both methods shows 10\% improvement in the overall relative loss. The relative L1-loss-Mask column shows the loss only for the ranges covered by the monocular method. In our experiment for depth < 1m}
  
  \begin{tabular}{ccc}
    \toprule
    Method&Rel. L1-loss&Rel. L1-loss-Mask\\
    \midrule
    Stereo & 0.071 & 0.172 \\
    Mono & 0.551 & 0.119\\
    Fused & \textbf{0.063} &  -  \\
  \bottomrule
  \label{table:fusion_loss}
\end{tabular}

\end{table}


\subsection{Limitations}
\label{limitations}
While showing improved results, there are still limitations that need to be considered when applying the suggested method: 
\begin{itemize}
    \item Once a phase-encoded mask is applied, the contrast of the acquired image decreases. Indeed, the image may be blindly deblurred using the knowledge of the PSF model, but the clear image reconstruction requires post-processing, as presented in \cite{Krishnan11Blind,EDOF_spr,EDOF_DL}. On the upside, the deblurred image exhibits an extended depth-of-field in this case.
    \item The phase-encoded depth estimation method shows superior results in a narrow range of depths, hence to achieve both the calibration and depth estimation improvement, the acquired images should include the ranges covered by the $\psi=-4..10$ parameter. Thus, if all the objects in the images are in a far range, the monocular depth estimation would not extract enough depth information to calibrate and improve the stereo cameras setup. If all are on a close range, the problem lies at the stereo depth estimation. However, the image-based monocular based method is not limited to a close range, hence can be applied to far scenes as demonstrated on KITTI dataset. 
\end{itemize}

\section{Conclusion} \label{sec:summ}

An approach for combining monocular depth cues and stereo disparity information is proposed. It allows avoiding the need for a costly and sensitive calibration process, and also to improve the overall depth estimation results. While the system is trained on simulated data, both of its features are examined in simulation as well as in real-world experiments. Note that no fine-tuning on real world images is done (after training on simulated images), which demonstrates the generalization and robustness of the system to various environments. 

An interesting direction for a future study is extending our technique to a sequence of images taken with a single camera from different points of view. In this case, our proposed auto-calibration feature will provide a fast and reliable method for calibrating each image pair, in order to produce a full 3D model of the captured scene \cite{Ummenhofer_2017_Demon}.

It is important to note that although the presented system provides both extended depth range and auto-calibration, each of these features can be achieved separately depending on the desired application. To use the phase-coded monocular method, One can equip only one of the stereo cameras with a phase mask, which will be used for either improving the depth accuracy or for auto-calibration. In this case, the clear aperture camera, which is used for the stereo depth, will provide a conventional high-contrast image.


This work presents a novel proof-of-concept for combining a general depth estimation technique with a monocular one to improve each other. We believe that this suggested solution opens the door to other combinations. For example, instead of the stereo camera in our system, one may use other passive or active depth estimation approaches (e.g., structured light). In the monocular side, notice that our scheme with only minor changes can employ other monocular depth estimation techniques instead of the two applied here.

\bibliographystyle{ACM-Reference-Format}
\bibliography{000_sample-sigconf}









\end{document}